\documentclass[10pt,twocolumn,letterpaper]{article}

\usepackage{afterpage}
\usepackage[pagenumbers]{cvpr}     
\usepackage[dvipsnames]{xcolor}

\definecolor{cvprblue}{rgb}{0.21,0.49,0.74}
\usepackage[pagebackref,breaklinks,colorlinks,citecolor=cvprblue]{hyperref}

\newcommand\mypara[1]{\vspace{1mm}\noindent\textbf{#1.}}

\def\paperID{12575}
\def\confName{CVPR}
\def\confYear{2024}

\title{EVP: Enhanced Visual Perception using Inverse Multi-Attentive Feature Refinement and Regularized Image-Text Alignment}

\author{Mykola Lavreniuk\\
SRI NASU-SSAU\\
\and
Shariq Farooq Bhat\\
KAUST\\
\and
Matthias M{\"u}ller\\
Intel Labs\\
\and
Peter Wonka\\
KAUST\\
}
\begin{document}
\maketitle

\begin{abstract}
This work presents the network architecture EVP (Enhanced Visual Perception). EVP builds on the previous work VPD which paved the way to use the Stable Diffusion network for computer vision tasks. We propose two major enhancements. First, we develop the Inverse Multi-Attentive Feature Refinement (IMAFR) module which enhances feature learning capabilities by aggregating spatial information from higher pyramid levels. Second, we propose a novel image-text alignment module for improved feature extraction of the Stable Diffusion backbone. The resulting architecture is suitable for a wide variety of tasks and we demonstrate its performance in the context of single-image depth estimation with a specialized decoder using classification-based bins and referring segmentation with an off-the-shelf decoder.
Comprehensive experiments conducted on established datasets show that EVP achieves state-of-the-art results in single-image depth estimation for indoor (NYU Depth v2, $11.8\%$ RMSE improvement over VPD) and outdoor (KITTI) environments, as well as referring segmentation (RefCOCO, $2.53$ IoU improvement over ReLA). The code and pre-trained models are publicly available at \href{https://github.com/Lavreniuk/EVP}{https://github.com/Lavreniuk/EVP}.

\end{abstract}

\begin{figure*}[t]
  \centering
   \includegraphics[width=1\linewidth]{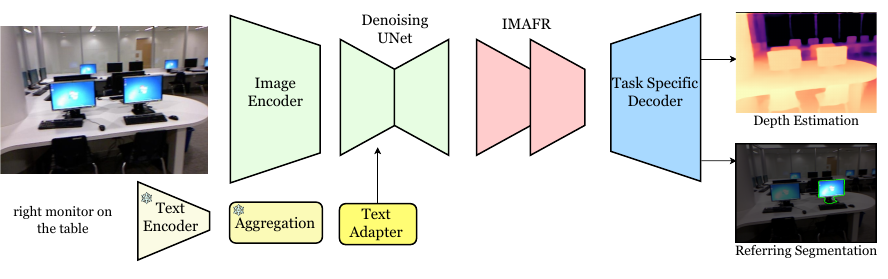}
   \hfill

   \caption{Overview of the EVP model architecture. An input image is first encoded by an auto-encoder and a denoising U-Net (light green) taken from a pre-trained Stable Diffusion model. Our proposed Inverse Multi-Attentive Feature Refinement (IMAFR) module (light red) refines features from the denoising U-Net at different scales. Our novel text aggregation strategy (yellow), combines information from class names or BLIP-2-generated captions to create a unified, enriched description for improved model performance.}
   \label{fig:our_model}
\end{figure*}

\section{Introduction}
Depth estimation is a core computer vision problem. Estimating depth is fundamental for many applications such as robotics (mapping, localization, planning, scene understanding, \etc), virtual reality, photography, and generative AI to name just a few. While there has been impactful work on relative depth estimation where per-pixel depth values are predicted up to some unknown scale, most applications require metric depth or at least benefit from it. 

This usually requires a calibrated stereo camera setup with a known baseline and camera parameters to triangulate corresponding pixels from both 2D planes and compute the metric depth. For many applications, it is desirable to predict depth from a single image, \eg, for single image editing, or in order to reduce system cost and complexity. While this problem is regarded as ill-posed, recent learning-based methods have achieved remarkable results in this setting.

A recent idea for depth estimation is to leverage recent progress in self-supervised learning. With the rise of large paired image-text datasets, self-supervised learning such as generative diffusion can extract information from a significant amount of data.
VPD demonstrated that the pre-trained U-Net backbone of Stable Diffusion can be leveraged for other computer vision tasks, such as depth estimation and referring segmentation. Due to the large-scale pre-training with text captions, this model generalizes well and contains a rich multi-modal context. 

In this work, we further improve VPD and expand it in two ways. First, we add our Inverse Multi-Attentive Feature Refinement (IMAFR) module which aggregates feature maps across the whole network using multi-attention. This provides more flexibility compared to more rigid hierarchical aggregation strategies. Second, we improve image-text alignment by using free-form text descriptions generated with vision-language models rather than relying on pre-defined object classes and description templates.

We evaluate EVP on two tasks, depth estimation and referring segmentation. For depth estimation, we also change the decoder, inspired by ZoeDepth, which further boosts performance. On both tasks, EVP outperforms current state-of-the-art methods. On the indoor depth benchmark NYU Depth v2, EVP reduces the RMSE by 11.8\% from 0.254 to 0.224 compared to the next best previous method VPD. EVP also establishes a new state-of-art on KITTI (outdoor depth) winning in all 7 metrics compared to the previous SOTA model GEDepth \cite{yang2023gedepth}. Finally, we achieve a new SOTA on RefCoco for referring segmentation improving the IoU by $2.53\%$ compared to ReLA.

In summary, our contributions are threefold: (1) We propose the novel Inverse Multi-Attentive Feature Refinement module for effective feature aggregation across layers, a regularized free-form image-text alignment module, and a classification-based decoder for depth estimation. (2) We integrate these modules with a Stable Diffusion backbone to form the novel network architecture EVP. (3) We conduct extensive experiments on depth estimation and referring segmentation outperforming current state-of-the-art methods.

\section{Related Work}
\textbf{Diffusion models}~\cite{rombach2022stablediffusion,ho2020denoising, nichol2021glide, nichol2021improved, dhariwal2021diffusion, ramesh2022hierarchical, balaji2022eDiff-I, saharia2022photorealistic} have recently demonstrated unprecedented success in image generation and have been subsequently adapted for a variety of tasks such as image inpainting~\cite{lugmayr2022repaint}, image-to-image translation tasks~\cite{brooks2023instructpix2pix} and even zero-shot video generation~\cite{khachatryan2023text2video,huang2023free} and 3D generation~\cite{poole2022dreamfusion}. Due to the requirement of input-output shape equality in the denoising process, a common theme while designing the architecture for diffusion models is to use a U-Net~\cite{Ronneberger2015u}. 
Latent diffusion models (LDMs)~\cite{rombach2022stablediffusion} first train an autoencoder~\cite{esser2021taming} and learn the denoising network in the latent space. Their denoising network's architecture is a derivative of the U-Net consisting of self- and cross-attention layers. In particular, latent diffusion shows that cross-attention layers can be employed for flexible conditioning via image or text features. Employing CLIP~\cite{radford2021clip} as a text encoder for cross-attention has led to the popular text-conditioned image generative LDM - Stable Diffusion. It quickly became evident that cross-attention maps between text and image features contain rich semantic information~\cite{hertz2022prompt}, hinting at the potential utility of generative text-to-image diffusion models for discriminatory vision tasks. 

\textbf{VPD}~\cite{zhao2023vpd} demonstrated that features learned by the denoising U-Net can indeed be exploited for vision tasks such as depth estimation and referring segmentation, outperforming prior works in both domains. VPD uses the cross-attention maps and the denoising U-Net's decoder features directly as input to a task-specific decoder whose architecture design is directly taken from the state-of-the-art architectures in the respective domains~\cite{yang2022lavt,xie2023swinv2lmim}. While the cross-attention maps provide rich semantic information, the features from the U-Net (trained for denoising) may not align well with the task semantics. In this work, we show that using our IMAFR module to align the features with the task semantics before feeding them to a task-specific decoder leads to a substantial improvement.

\textbf{Depth estimation} has seen significant advancements along two major fronts: reformulation and leveraging pre-training techniques. AdaBins~\cite{bhat2021adabins} and the subsequent adaptive bin-based methods~\cite{bhat2022localbins,li2022binsformer,pixelbinsSarwari:EECS-2021-32,bhat2023zoedepth,agarwal2022attention,shao2023iebins,lee2023slabins} reformulate depth estimation as a classification-regression task adaptively discretizing the depth interval into bins and subsequently representing depth as a linear combination of bin centers and corresponding predicted probabilities. ZoeDepth~\cite{bhat2023zoedepth} showed that large relative depth pre-training results in significant improvements. On the other hand, \cite{xie2023swinv2lmim} demonstrated large-scale pre-training via masked image modeling (MIM) also leads to state-of-the-art performance. Finally, large scale training of text-to-image diffusion models can also be considered as a form of pre-training and truly leads to remarkable performance improvements~\cite{zhao2023vpd}. Our work directly builds on and improves this current state-of-the-art model. 

\textbf{Referring segmentation} aims to precisely locate objects at the pixel level within an image based on a provided referring expression. Previous research focused on two key aspects: (1) the extraction of features from both visual and language domains, and (2) the fusion of these multi-modal features. Numerous methodologies have been explored for feature extraction, ranging from the application of CNNs \cite{Chen_2019_ICCV, Hu_2020_CVPR, Huang_2020_CVPR, yu2018mattnet} and recurrent neural networks \cite{yu2018mattnet, Feng_2021_CVPR} to transformer models \cite{kim2022restr, yang2022lavt, ding2022vlt, zhao2023vpd, liu2023rela, Liu_2023_CVPR, Yan_2023_CVPR}. Recent papers \cite{Liu_2023_CVPR, Yan_2023_CVPR} utilized additional datasets for pre-training their models before training on RefCOCO. Therefore, we exclude these two papers from direct comparison with our model to ensure fair evaluation.

We also acknowledge concurrent work on arXiv with good results in single image depth estimation~\cite{yang2023polymax,khan2023mesa,kondapaneni2023textimage,wang2023sqldepth}. Specifically, the work~\cite{kondapaneni2023textimage} also aims to improve image-text alignment. 

\section{Methodology}
In this section, we provide an in-depth exposition of our architecture, discuss our design decisions, and outline the specifics of our training protocol.

\subsection{Preliminaries}
\mypara{Stable Diffusion}
Our model is built on the popular Stable Diffusion model~\cite{rombach2022stablediffusion} which is trained on the extensive LAION-5B image-text dataset. It comprises four key components: an encoder denoted as $E$, a conditional denoising autoencoder with a U-Net structure represented as $\epsilon_{\theta}$, a language encoder $\tau_{\theta}$ utilizing the CLIP~\cite{radford2021clip} text encoder, and a decoder $D$.
The autoencoder is trained with a combination of losses to ensure accurate and realistic reconstructions. Specifically, it integrates a perceptual loss and a patch-based adversarial objective.
Moreover, in the pre-training phase, both the encoder $E$ and the decoder $D$ are trained before the denoising autoencoder $\epsilon_{\theta}$, establishing the condition $D(E(x)) = \tilde{x} \approx x$. This strategy ensures robust reconstructions within the image manifold. Subsequently, the diffusion model is trained in this latent space, guided by the objective:
\[
L_{LDM} := E_{E(x), y, \epsilon \sim \mathcal{N}(0,1), t} \left\| \epsilon - \epsilon_{\theta}(z_t, t, \tau_{\theta}(y)) \right\|^2_2
\]
where \( z_t \) is the latent representation, that can be efficiently obtained from $E$ during training and \( t \) is the time step.

\mypara{Visual Perception with a pre-trained Diffusion model (VPD)}
The Visual Perceptual Diffusion (VPD)~\cite{zhao2023vpd} model builds on a pre-trained diffusion model. VPD takes advantage of the rich, high-level context embedded in the text captions used during pre-training by providing text descriptions or prompts for input images. The prediction model is redefined as $p_{\phi}(y|x, S)$, where $x$ is the input image and $S$ represents the set of relevant text descriptions or prompts associated with the input image $x$. For referring segmentation a text prompt is provided already. For depth estimation, VPD uses category name labels to generate various captions. For example, $S$ could be a set of 80 captions that are created by applying text templates, such as "a bad photo of a \{\}", to a room name, such as "bathroom". Therefore, the framework requires a text label describing each input image.
The formulation involves three key components:

$p_{\phi1}(C|S)$ extracts text features from the generated captions or provided prompts, utilizing a CLIP text encoder from the pre-training stage of Stable-Diffusion and a text adapter -- a refinement step with a two-layer MLP.

$p_{\phi2}(F|x, C)$ extracts hierarchical feature maps based on the input image and conditioned on the text features. The pre-trained text-to-image diffusion model serves as an excellent initialization for this process.

$p_{\phi3}(y|F)$ is a lightweight prediction head generating results from the hierarchical feature maps.

The final prediction is calculated as:
\[p_{\phi}(y|x, S) = p_{\phi_3}(y|F) p_{\phi_2}(F|x, C) p_{\phi_1}(C|S) \quad \]

\subsection{Overview}
The VPD architecture utilizes the image-text cross-attention maps and the denoising U-Net's decoder features as input to the task-specific decoder. While cross-attention maps provide a powerful prior, we conjecture that the U-Net features, initially trained for noise prediction, do not align well with the semantics of the task (depth estimation or referring segmentation). This leaves an undue burden on the task-specific decoder which is often rather lightweight~\cite{yang2022lavt,xie2023swinv2lmim}. 
To this end, we propose a novel encoder block designed to encode the U-Net decoder features -- IMAFR (Sec.~\ref{sec:IMAFR}). Additionally, 
we automatically generate rich textual descriptions of the input image instead of template descriptions that rely on category names as used by VPD and propose a novel aggregation strategy (Sec.~\ref{sec:TextAlignment}). Finally, we propose a decoder specifically for depth estimation (Sec.~\ref{sec:DepthDecoder}).

\subsection{Inverse Multi-Attentive Feature Refinement (IMAFR)}
\label{sec:IMAFR}

\begin{figure}[t]
  \centering
   \includegraphics[width=1\linewidth]{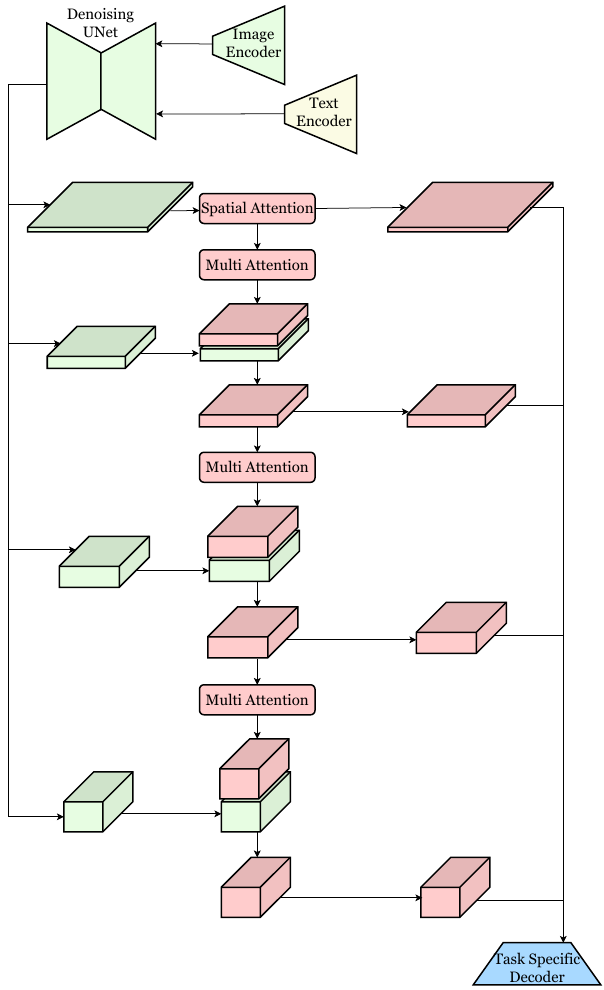}
   \hfill
   \caption{Inverse Multi-Attentive Feature Refinement (IMAFR) (light pink) adeptly refines features at different scales received from the denoising U-Net (light green) using multi-attention.}
   \label{fig:our_block}
\end{figure}

Our novel Inverse Multi-Attentive Feature Refinement (IMAFR) shown in \cref{fig:our_model} diverges from the well-known pyramid aggregation used by FPN and U-Net~\cite{lin2017fpn, Ronneberger2015u}, which relies on the top-down pathway to enrich higher resolution features through upsampling spatially coarser, but semantically stronger feature maps from hierarchical pyramid levels. In contrast, inspired by previous studies on pyramid feature aggregation~\cite{Zhou2019unet++, Zhou2018unet++, Liu2018CU-Net, Gubins2020Cascaded-UNet}, our method prioritizes spatial information within feature maps from higher pyramid levels. These maps, rich in spatial details, are particularly valuable for tasks like monocular depth estimation and referring segmentation where the output is a dense image prediction rather than a class since the importance of higher resolution features is emphasized.

The IMAFR module introduces a new approach to feature refinement using a multi-attention mechanism to enhance features from different scales. This mechanism is inspired by previous studies on different attention blocks~\cite{Hu_2018_CVPR, Li_2019_CVPR, Wang_2020_CVPR, Woo2018CBAM} and incorporates spatial attention, channel attention, and group normalization, collectively contributing to refined feature extraction.
The hierarchical features $p_{\phi2}(F|x, C)$, extracted using the Diffusion U-Net, serve as input to our IMAFR block \cref{fig:our_block}, enhancing the refinement process with an additional component $p_{\phi_4}(F_e|F)$. IMAFR ensures that the most important details are kept and adds valuable information from higher pyramid levels.

The prediction model is now calculated as:
\[ p_{\phi}(y|x, S) = p_{\phi_3}(y|F_e) p_{\phi_4}(F_e|F) p_{\phi_2}(F|x, C) p_{\phi_1}(C|S), \quad \]
where set of features with different scales is represented by $F = \{f_1, f_2, f_3, f_4\}$ and $F_e = \{fe_1, fe_2, fe_3, fe_4\}$. Here, 

\begin{align*}
& fe_i = \text{Conv}\left(\text{Concat}\left(\text{MultiAttention}(fe_{i-1}), f_i\right)\right), i \in [1,3] \\
& fe_4 = \text{{SpatialAttention}}(f_4)
\end{align*}

where $\text{Conv}$ represents the module that includes a 2D convolution operation with a 1x1 kernel, GroupNorm, and ReLU activation function and $\text{MultiAttention}$ block that successively applies spatial attention, followed by channel attention, and then two consecutive $\text{Conv}$ blocks.

We also normalize the latent space based on the component-wise standard deviation to reduces the signal-to-noise ratio \cite{rombach2022stablediffusion}. Consequently, we compute the component-wise standard deviation value of the encoder latent space $std$ across the entire dataset as a pre-processing step. Hence, we replace the module $p_{\phi2}(F|x, C)$ that extracts hierarchical feature maps based on the input image and conditions on the text features with
$p_{\phi2}(F|x, std, C)$.

\begin{figure*}[t]
  \centering
   \includegraphics[width=1\linewidth]{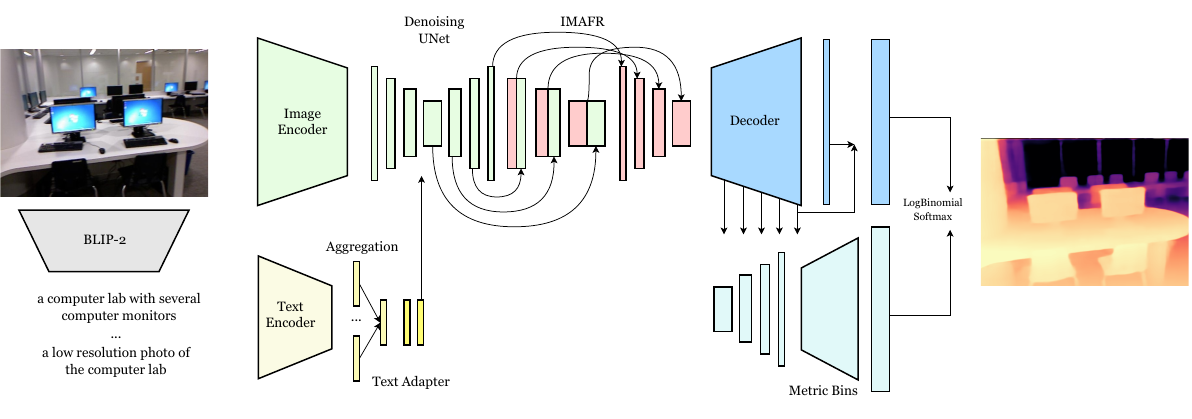}
   \hfill
   \caption{Detailed illustration of the EVP model architecture. Each component, including the Inverse Multi-Attentive Feature Refinement (IMAFR) module (light pink) and the novel text aggregation strategy (warm yellow), provides a comprehensive view of the model's internal structure and information flow. The IMAFR module adeptly refines features at various scales, leveraging critical spatial information from higher pyramid levels. The novel text aggregation strategy combines information from class names or BLIP-2-generated captions (light gray), creating a unified, enriched description to enhance overall model performance.}
   \label{fig:our_model_depth}
\end{figure*}

\subsection{Regularized Image-Text Alignment}
\label{sec:TextAlignment}

High-level knowledge and context embedded in natural language descriptions and low-level image features are complementary and fusing them leads to increased robustness and generalization performance. Training multi-modal models with aligned vision and language features enables zero-shot transfer to many different applications \cite{radford2021clip}.  Recently, works like VPD \cite{zhao2023vpd} have shown that this insight also transfers to dense prediction tasks. In VPD, image captions are generated by populating predefined text description templates with category names (\eg, room names). These captions are then embedded with CLIP and used to guide the image features extracted with the U-Net from a pre-trained diffusion model. However, most datasets do not provide such explicit labels which is likely why VPD did focus on generating results for NYUv2, as this dataset has room names as labels. While templates could probably be devised to leverage object class detections or meta-data such as capture location, this design choice is not very scalable and also requires each image to be annotated at test time. 

To overcome both of these challenges, we introduce a novel approach to image-text alignment. First, we automatically generate free-form image captions leveraging advanced models like BLIP-2 \cite{li2023blip2}. This approach can generate more specific descriptions and is also more scalable. Hence, we generate descriptions for the complete dataset and embed them with CLIP \cite{radford2021clip} before training. The best results can be achieved when using 40 CLIP vectors of size $768\times1$ to describe each image.

However, this type of guidance may be too specific or noisy making training more challenging (\eg, easier to overfit or underfit) and still requires captioning at test time. As an alternative, we can aggregate all embeddings across the dataset to obtain a single set of 40 text embedding vector for the complete dataset: \[
p_{\phi1}(C|S) = \frac{1}{|S|} \sum_{s \in S} p_{\phi1}(C|s),
\]

An aggregated set of embedding vectors represents a rich summary of the domain and can be used both during training and testing. Surprisingly, this works almost as well as using image-specific embeddings and has the advantage that it is not necessary to generate text embeddings for new images during test time.

Finally, we explore many alternative approaches to image-text alignment and compare them in the ablation study in \ref{sec:ablation}. 

\subsection{Specialized Decoder for Depth Estimation}
\label{sec:DepthDecoder}

The previously described model utilizing the Inverse Multi-Attentive Feature Refinement (IMAFR) module and Regularized Image-Text Alignment (RITA) is primarily designed to excel in both referring segmentation and monocular depth estimation tasks. While it offers promising results in both domains, recent research has highlighted a novel approach to depth estimation. It has been demonstrated that treating depth estimation as a classification task can lead to more accurate results compared to regression-based methods. To utilize the benefits of classification-based depth estimation, we have extended the model's decoder with components inspired by the ZoeDepth model \cite{bhat2023zoedepth}. The incorporation of these depth-specific components enhances the accuracy of depth estimation, offering a performance boost. The final architecture for monocular depth estimation, featuring a decoder design inspired by ZoeDepth, is depicted in \cref{fig:our_model_depth}. This configuration is created specifically for depth estimation, ensuring the model excels in this task.

\begin{table}
\centering
\small
\begin{tabular}{l@{\hspace{4pt}}c@{\hspace{4pt}}c@{\hspace{6pt}}c@{\hspace{6pt}}c@{\hspace{5pt}}c@{\hspace{5pt}}c@{\hspace{4pt}}}
\toprule
Method & RMSE↓ & $\delta_1$ ↑ & $\delta_2$ ↑ & $\delta_3$ ↑ & REL ↓ & $\log_{10}$ ↓ \\
\hline
BTS \cite{bts_lee2019big} & 0.392 & 0.885 & 0.978 & 0.995 & 0.110 & 0.047 \\
AdaBins \cite{bhat2021adabins} & 0.364 & 0.903 & 0.984 & 0.997 & 0.103 & 0.044 \\
DPT \cite{Ranftl_2021_ICCV_DPT} & 0.357 & 0.904 & 0.988 & 0.998 & 0.110 & 0.045 \\
P3Depth \cite{patil2022p3depth} & 0.356 & 0.898 & 0.981 & 0.996 & 0.104 & 0.043 \\
NeWCRFs \cite{yuan2022new} & 0.334 & 0.922 & 0.992 & 0.998 & 0.095 & 0.041 \\
SwinV2-B \cite{liu2022swin} & 0.303 & 0.938 & 0.992 & 0.998 & 0.086 & 0.037 \\
SwinV2-L \cite{liu2022swin} & 0.287 & 0.949 & 0.994 & 0.999 & 0.083 & 0.035 \\
AiT \cite{ning2023ait} & 0.275 & 0.954 & 0.994 & 0.999 & 0.076 & 0.033 \\
ZoeDepth \cite{bhat2023zoedepth} & 0.270 & 0.955 & 0.995 & 0.999 & 0.075 & 0.032 \\
VPD \cite{zhao2023vpd} & 0.254 & 0.964 & 0.995 & 0.999 & 0.069 & 0.030 \\
\toprule
\textbf{EVP} & \textbf{0.224} & \textbf{0.976} & \textbf{0.997} & \textbf{0.999} & \textbf{0.061} & \textbf{0.027} \\
\bottomrule
\end{tabular}
\caption{Performance comparison on the NYU Depth v2 dataset. The provided values are sourced from the respective original papers. The best results are highlighted in bold.}
\label{tab:nyu_results}
\end{table}

\begin{table*}
\centering
\begin{tabular}{@{}lccccccc@{}}
\toprule
Method & REL↓ & SqREL↓ & RMSE↓ & RMSE log↓ & $\delta_1$ ↑ & $\delta_2$ ↑ & $\delta_3$ ↑ \\
\hline
BTS \cite{bts_lee2019big} & 0.061 & 0.261 & 2.834 & 0.099 & 0.954 & 0.992 & 0.998 \\
AdaBins \cite{bhat2021adabins} & 0.058 & 0.190 & 2.360 & 0.088 & 0.964 & 0.995 & \underline{0.999} \\
ZoeDepth \cite{bhat2023zoedepth} & 0.057 & 0.194 & 2.290 & 0.091 & 0.967 & 0.995 & \underline{0.999} \\
NeWCRFs \cite{yuan2022new} & 0.052 & 0.155 & 2.129 & 0.079 & 0.974 & \underline{0.997} & \underline{0.999} \\
iDisc \cite{piccinelli2023idisc} & \underline{0.050} & 0.148 & 2.072 & 0.076 & 0.975 & \underline{0.997} & \underline{0.999} \\
NDDepth \cite{shao2023nddepth} & \underline{0.050} & 0.141 & 2.025 & \underline{0.075} & \underline{0.978} & \textbf{0.998} & \underline{0.999} \\
SwinV2-L 1K-MIM \cite{xie2023swinv2lmim} & \underline{0.050} & \underline{0.139} & \textbf{1.966} & \underline{0.075} & 0.977 & \textbf{0.998} & \textbf{1.000} \\
GEDepth \cite{yang2023gedepth} & \textbf{0.048} & 0.142 & 2.044 & 0.076 & 0.976 & \underline{0.997} & \underline{0.999} \\
\toprule
\textbf{EVP} & \textbf{0.048} & \textbf{0.136} & \underline{2.015} & \textbf{0.073} & \textbf{0.980} & \textbf{0.998} & \textbf{1.000} \\
\bottomrule
\end{tabular}
\caption{Performance comparison on the KITTI dataset for single frame methods. The provided values are sourced from the respective original papers. The best results are highlighted in bold, second best are underlined.}
\label{tab:kitti_results}
\end{table*}

\section{Results}
In the following, we present comprehensive experimental results, providing empirical evidence for the effectiveness of our proposed approach.
We report results on well-established datasets for single-image depth estimation in both indoor (NYU Depth v2) and outdoor (KITTI) environments, as well as referring segmentation (RefCOCO). We first provide an overview of these datasets and the evaluation metrics employed. Then, we present quantitative comparisons against previously published state-of-the-art models and ablation studies.

\subsection{Datasets} 

{\bf NYU Depth v2} comprises images and corresponding depth maps captured in various indoor scenes, all at a pixel resolution of 640 × 480. This dataset encompasses 120,000 training samples and 654 testing samples. Our training process utilizes a subset of 50,000 samples. Notably, the depth maps have a maximum range of 10 meters.

{\bf KITTI} presents a collection of outdoor scenes, captured from a car equipped with stereo imaging and 3D laser scanning technology. The RGB images exhibit a resolution of roughly 1241 × 376 pixels.
During training, our network utilizes a subset of approximately 26,000 left-view images, excluding scenes featured in the 697-image test set. The depth maps in this dataset are constrained by a maximum range of 80 meters.

{\bf RefCOCO} includes roughly 20,000 images and 50,000 annotated objects, along with a vast collection of 142,209 expressions. In accordance with standard convention, we train our model using the training set and evaluate it on the validation set.

\subsection{Metrics}

We use the standard metrics for depth estimation, which include the absolute relative error (REL), root mean squared error (RMSE), RMSE log, squared relative difference (Sq. REL), average $\log_{10}$ error between predicted depth $\hat{d}$ and the ground truth depth $d$, the threshold accuracy $\delta_n$, which is defined as $\delta_n = \%$ of pixels satisfying $\max\left(\frac{d_i}{\hat{d}_i}, \frac{\hat{d}_i}{d_i}\right) < 1.25^n$ for $n = 1, 2, 3$. See ~\cite{Eigen2014} for an explanation of these metrics.
We use the standard metric of overall intersection-over-union (IoU) for referring segmentation~\cite{yu2018mattnet}.

\subsection{Depth Estimation}
We compare our method to the current published state-of-the-art methods for single image metric monocular depth estimation on two datasets, NYUv2 and KITTI. We consider VPD as our main competitor for NYUv2 and SwinV2-L 1K-MIM as well as GEDepth as our main competitors for KITTI.
The results for NYUv2 are shown in Table~\ref{tab:nyu_results}. Our method EVP beats the currently best method VPD in all metrics by a large margin. Our method improves both the REL and RMSE metrics by over 10\%, which is significant. For example, the RMSE improvements achieved by the previous two state-of-the-art methods were 5.9\% and 1.8\%, respectively. The results for KITTI are shown in Table~\ref{tab:kitti_results}. Our EVP model establishes a new state-of-art winning in all 7 metrics compared to the previous SOTA model GEDepth \cite{yang2023gedepth}.

\begin{figure*}[t]
  \centering

  \begin{subfigure}{\textwidth}
    \includegraphics[width=\linewidth]{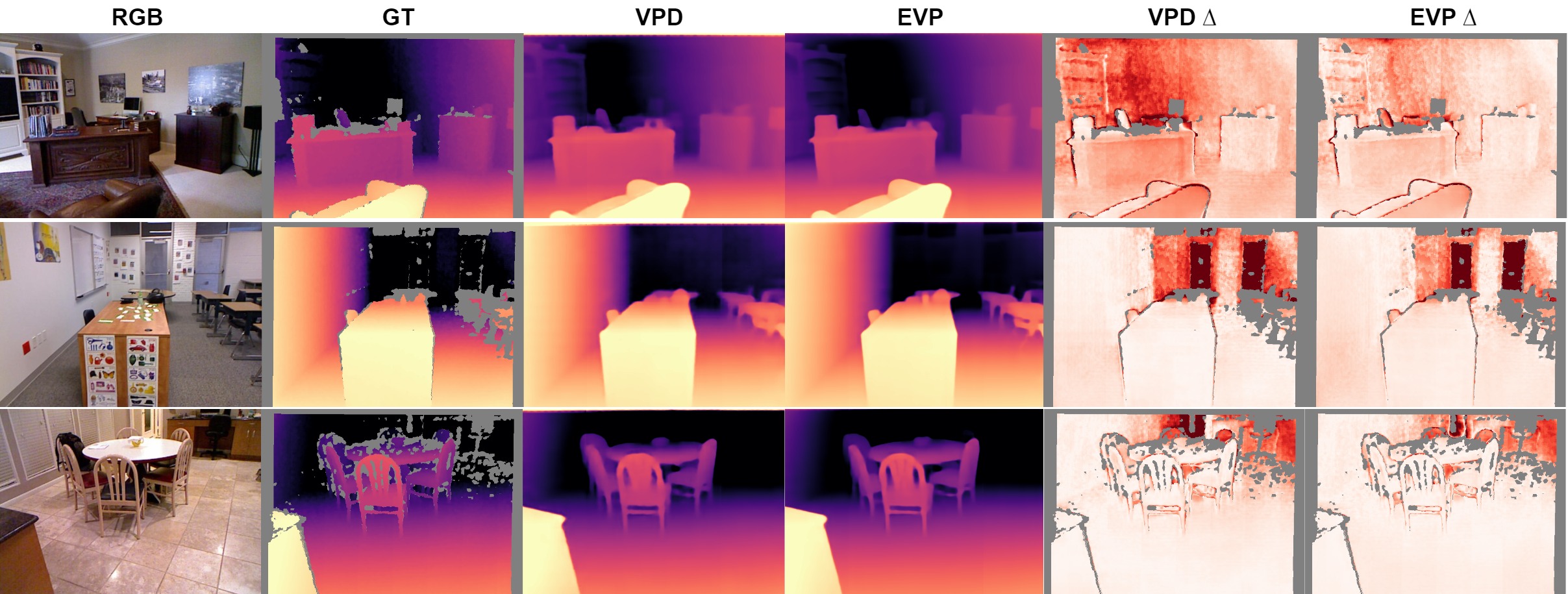}
    \caption{Visualization of EVP on images from the NYU Depth v2 dataset.}
    \label{fig:subfig4a}
  \end{subfigure}

  \begin{subfigure}{\textwidth}
    \includegraphics[width=\linewidth]{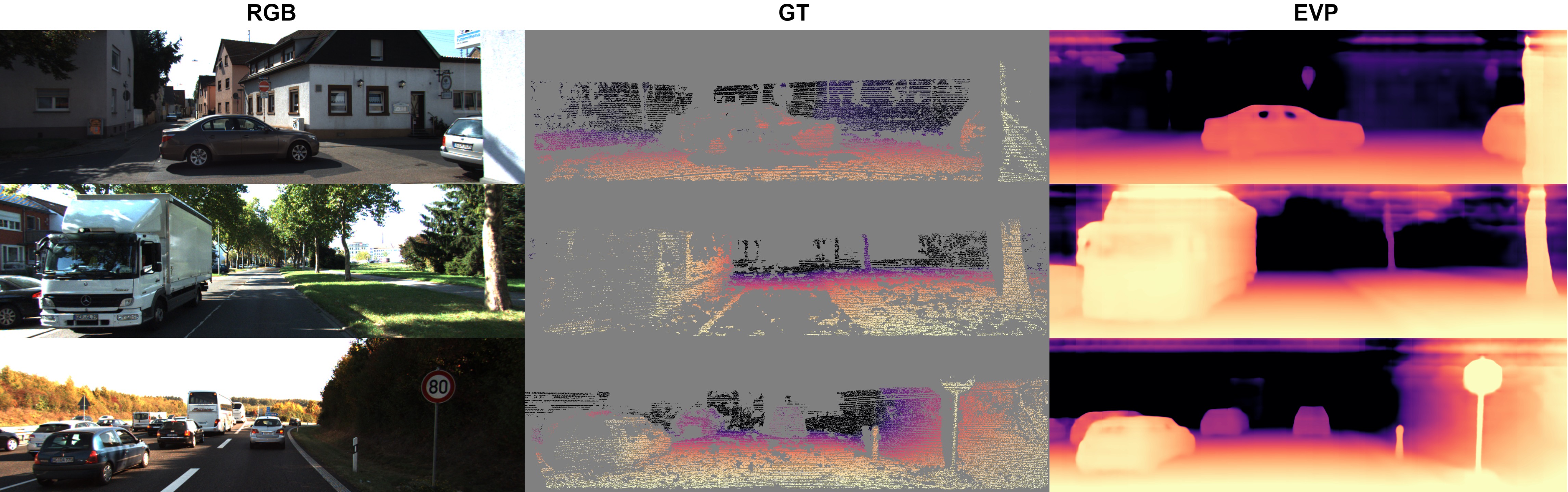}
    \caption{Visualization of EVP on images from the KITTI dataset.}
    \label{fig:subfig4b}
  \end{subfigure}

  \begin{subfigure}{\textwidth}
    \includegraphics[width=\linewidth]{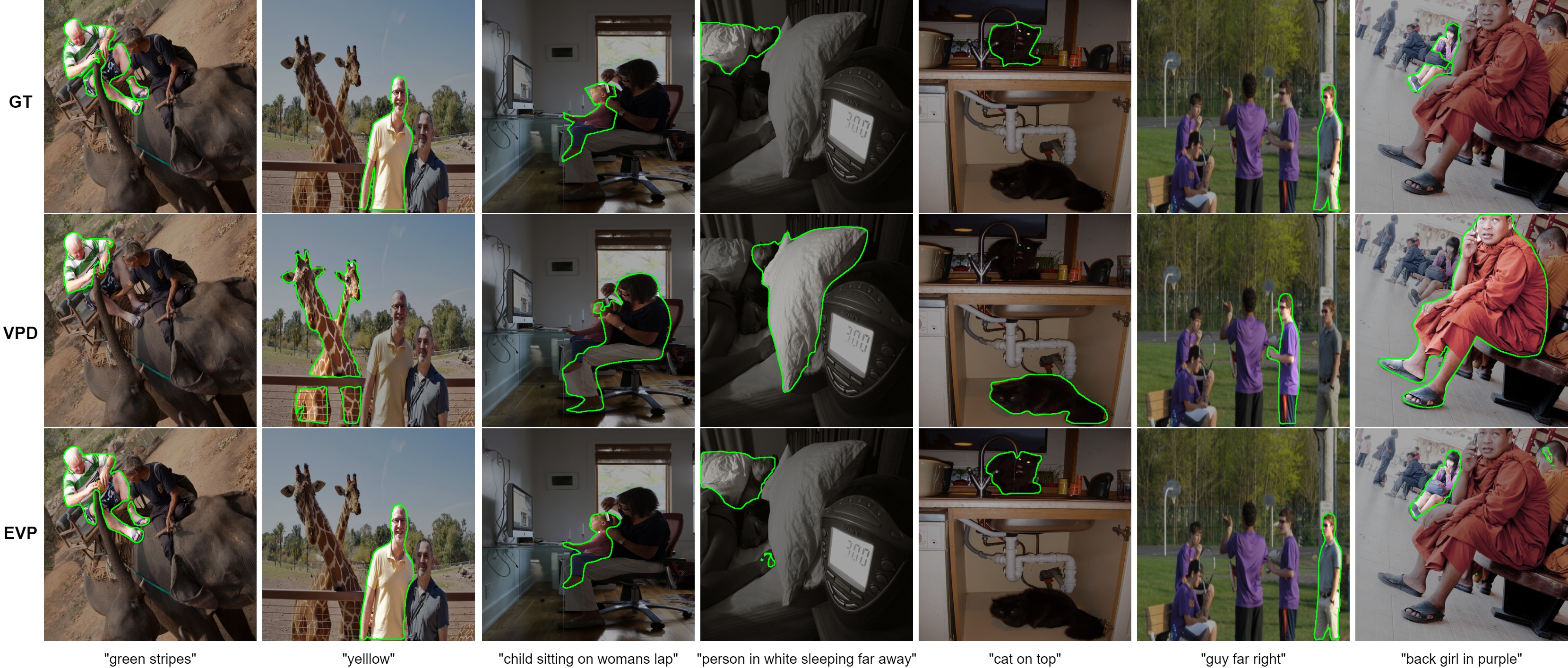}
    \caption{Visualization of EVP on images from the RefCOCO dataset.}
    \label{fig:subfig4c}
  \end{subfigure}

  \caption{Qualitative results of EVP on indoor and outdoor monodepth estimation and referring segmentation.}
  \label{fig:figure4}
\end{figure*}

\cref{fig:subfig4a} and ~\cref{fig:subfig4b} show visualizations of selected results on NYUv2 and KITTI respectively. We include error map visualizations to understand the types of errors made by models. We observe that VPD produces significant errors at large depth ranges, whereas our model diminishes the large range errors significantly. 
Surprisingly, for KITTI, we observe that our model is able to perform well even on thin objects, for example, pole signs, even though the native resolution supported by our Stable Diffusion backbone is small and no skip-connections with resolutions higher than $64\times64$ are available. We attribute this to our high-resolution refinement by IMAFR.

\subsection{Referring Segmentation}
We compare our method to the current published state-of-the-art methods trained only on the RefCOCO dataset. VPD and ReLA are our main competitors. \cref{tab:refcoco_results} lists the results in terms of the overall IoU metric on the RefCOCO dataset. Our proposed EVP architecture outperforms our baseline VPD as well as the current state-of-the-art ReLA model yielding a significant improvement of $+2.53$ IoU. 
This improvement is significantly higher than the current trend (+0.57 and +0.29 for the prior two works, respectively)  
See~\cref{fig:subfig4c} for a visualization of example results.

\subsection{Ablation Study}
\label{sec:ablation}
The ablation study shows the contributions of specific components within the EVP model, clarifying the respective impact on visual perception tasks. We use depth estimation on the NYU-Depth-v2 dataset for our ablation study. We report the results in \cref{tab:nyu_ablation} and explain them below.

The result in row 1 represents VPD. Adding only the IMAFR module demonstrates a substantial increase in accuracy by effectively leveraging hierarchical image features (row 2). We observe further improvements when adding the metric bins module and normalizing the latent space by the component-wise standard deviation (rows 3 and 4). Additionally, our regularized image-text alignment module (rows 10-12) also leads to a significant accuracy enhancement. We have explored several variations for image-text alignment and we briefly describe each alternative in the following. 

Directly using BLIPv2 descriptions per image (row 5) does not perform well. We conjecture that the rich individual descriptions for each image make learning more difficult and are more prone to noise; some level of abstraction seems to be beneficial, especially when using CLIP with pre-trained weights. We find that this can be overcome to a large extent by fine-tuning the CLIP weights (row 6). Using template descriptions based on the room label per image category as proposed by VPD works well in conjunction with our proposed modules (row 7). However, obtaining these category labels is not scalable and such annotation may not be available during test time or even during training time. Hence, our approach uses a single aggregated embedding based on per-image descriptions automatically generated by BLIPv2. Our approach performs best while not requiring explicit class labels. Rows 8 and 9 show the impact of only removing the IMFAR or metric bins module from our final architecture. Row 10 shows the result when computing a single $768 \times 1$ CLIP vector for the complete dataset. Row 11 shows the result for computing a set of 40 averaged $768 \times 1$ CLIP vectors for the complete dataset. Finally, row 12 shows our best method with 40 CLIP vectors extracted per image.

\begin{table}
\centering
\small
\begin{tabular}{l@{\hspace{4pt}}c@{\hspace{5pt}}c@{\hspace{7pt}}c@{\hspace{6pt}}}
\toprule
Method & Visual
Encoder & Textual
Encoder & overall IoU ↑ \\
\hline
MCN \cite{luo2020mcn} & Darknet53 & bi-GRU & 62.44 \\
ReSTR \cite{kim2022restr} & ViT-B  & Transformer & 67.22 \\
VLT \cite{ding2022vlt} & Darknet53 & bi-GRU & 67.52 \\
CRIS \cite{wang2022cris} & CLIP-R101 & CLIP & 70.47 \\
LAVT \cite{yang2022lavt} & Swin-B & BERT & 72.73 \\
VLT \cite{ding2022vlt} & Swin-B & BERT & 72.96 \\
VPD \cite{zhao2023vpd} & Stable Diffusion & CLIP & 73.25 \\
ReLA \cite{liu2023rela} & Swin-B & BERT & 73.82 \\
\toprule
\textbf{EVP} & Stable Diffusion & CLIP & \textbf{76.35} \\
\bottomrule
\end{tabular}
\caption{Performance comparison on the RefCOCO dataset. The provided values are sourced from the respective original papers. The best results are highlighted in bold.}
\label{tab:refcoco_results}
\end{table}

\begin{table}
\centering
\begin{tabular}{c@{\hspace{4pt}} c@{\hspace{4pt}}c@{\hspace{4pt}}c@{\hspace{4pt}}c@{\hspace{4pt}}c@{\hspace{4pt}}c@{\hspace{4pt}}c@{\hspace{4pt}}c@{\hspace{4pt}}c@{\hspace{4pt}}}
\toprule
ID & IMAFR & Bins & STD & ITA & Reg & CLIP & RMSE↓ & REL ↓ \\
\midrule
1 &- & - & - & cd & v & \checkmark & 0.254 & 0.069 \\
2 & \checkmark & - & - & cd & v & \checkmark & 0.243 & 0.066 \\
3 & \checkmark & \checkmark & - & cd & v & \checkmark & 0.242 & 0.066 \\
4 & \checkmark & \checkmark & \checkmark & cd & v & \checkmark & 0.238 & 0.065 \\
5 & \checkmark & \checkmark & \checkmark & id & v & \checkmark & 0.263 & 0.073 \\
6 & \checkmark & \checkmark & \checkmark & id & v & - & 0.229 & 0.062 \\
7 & \checkmark & \checkmark & \checkmark & cd & vd & \checkmark & 0.228 & 0.063 \\
8 & - & \checkmark & \checkmark & id & vd & \checkmark & 0.234 & 0.064 \\
9 & \checkmark & - & \checkmark & id & vd & \checkmark & 0.228 & 0.063 \\
10 & \checkmark & \checkmark & \checkmark & id & vd & \checkmark & 0.227 & 0.062 \\
11 & \checkmark & \checkmark & \checkmark & id & d & \checkmark & 0.226 & 0.062 \\
\midrule
12 & \checkmark & \checkmark & \checkmark & id & i & \checkmark & \textbf{0.224} & \textbf{0.061} \\

\bottomrule
\end{tabular}
\caption{Comparison of different design choices for EVP for Monocular Depth Estimation on the NYU-Depth-v2 dataset. Bins: metric bins module is used in the decoder, STD: latent space was divided by the component-wise standard deviation, \(ITA\): Image-Text Alignment using class description (cd) generated by substituting the room name into ImageNet templates or free-form image-level description (id) generated by BLIPv2 captioning model, Reg: if single regularized embedding across CLIP vectors (v), across dataset (d), across CLIP vectors and all dataset (vd); i - individual embedding, CLIP: frozen CLIP weights during EVP training.}
\label{tab:nyu_ablation}
\end{table}

\section{Conclusion}
In this paper, we have proposed a new model called Enhanced Visual Perception (EVP), which significantly improves upon the state-of-the-art in two computer vision tasks. Through the integration of our novel Inverse Multi-Attentive Feature Refinement (IMAFR) and Regularized Image-Text Alignment (RITA) modules, EVP excels in tasks like monocular depth estimation and referring segmentation.
EVP outperforms current state-of-the-art methods for monocular metric depth estimation on NYU Depth v2 and on KITTI. It also excels at referring segmentation, setting a new state-of-the-art on RefCOCO. 

We also discovered two limitations of our work. First, while we have the best overall performance in metric depth estimation, we inherit a limitation of VPD -- the boundaries of depth predictions are not as sharp as some other methods. Second, the number of parameters of the model is large (close to 1B), due to using the SD U-Net. This makes it hard to use the model on edge devices.

An interesting avenue for future work is exploring EVP's potential for different applications, such as semantic segmentation, instance segmentation, and object detection. We also think a tighter coupling between depth estimation and depth-conditioned image generation would be interesting, \eg, developing a single model that can predict the depth of an input image, generate RGBD images from scratch, and generate an image from depth-conditioning information. We hope this work will inspire further advances leveraging priors from large-scale data for computer vision tasks.

{
    \small
    \bibliographystyle{ieeenat_fullname}
    \bibliography{arxiv}
}

\clearpage
\newpage

\begin{minipage}{\textwidth}
  \centering
   \includegraphics[width=1\linewidth]{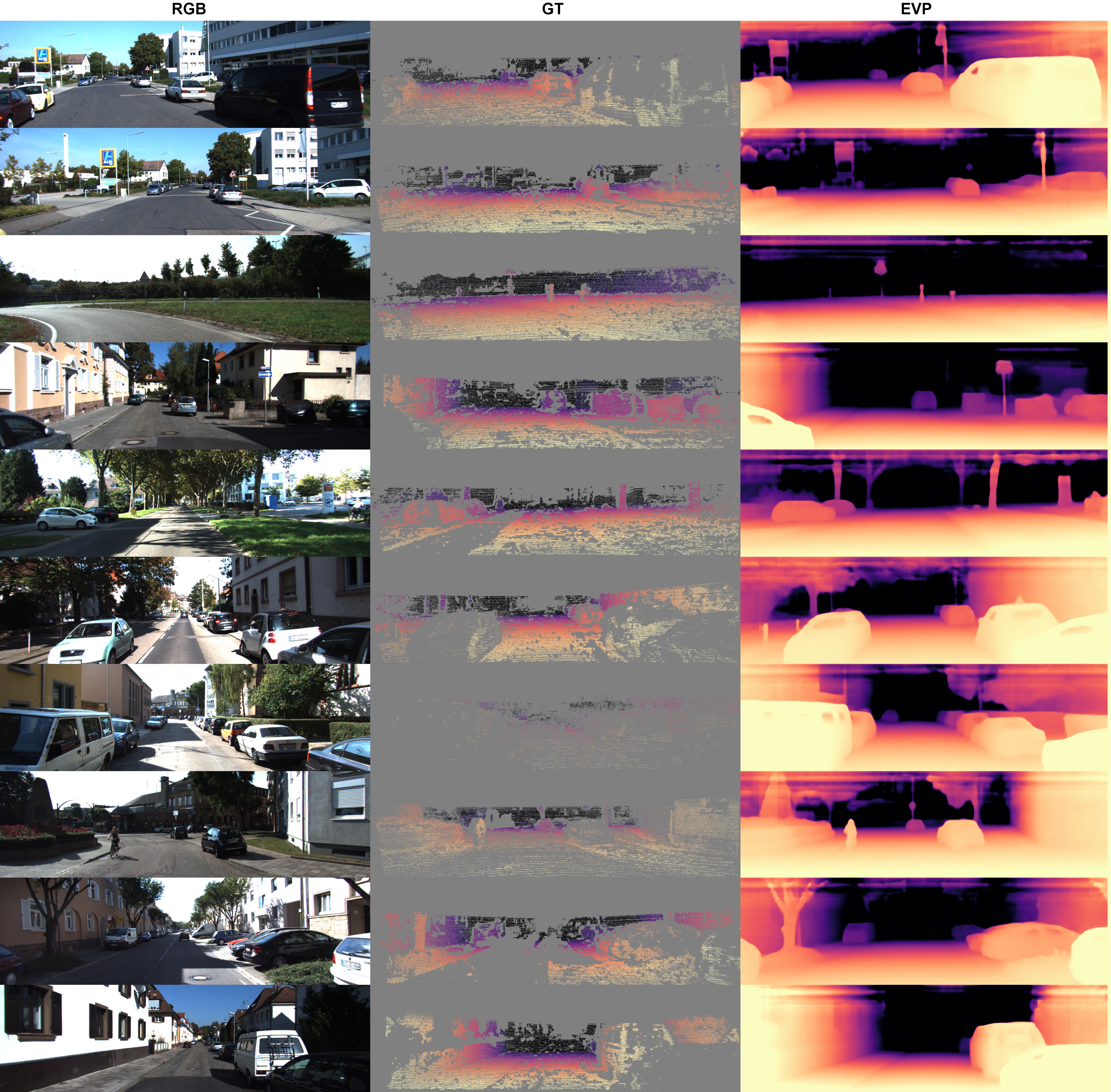}
   \captionof{figure}{Visualization of EVP on images from the KITTI dataset.}
   \label{fig:supsubfig2}
\end{minipage}

\begin{figure*}[t]
  \hspace{0.5cm}
   \includegraphics[width=1\linewidth]{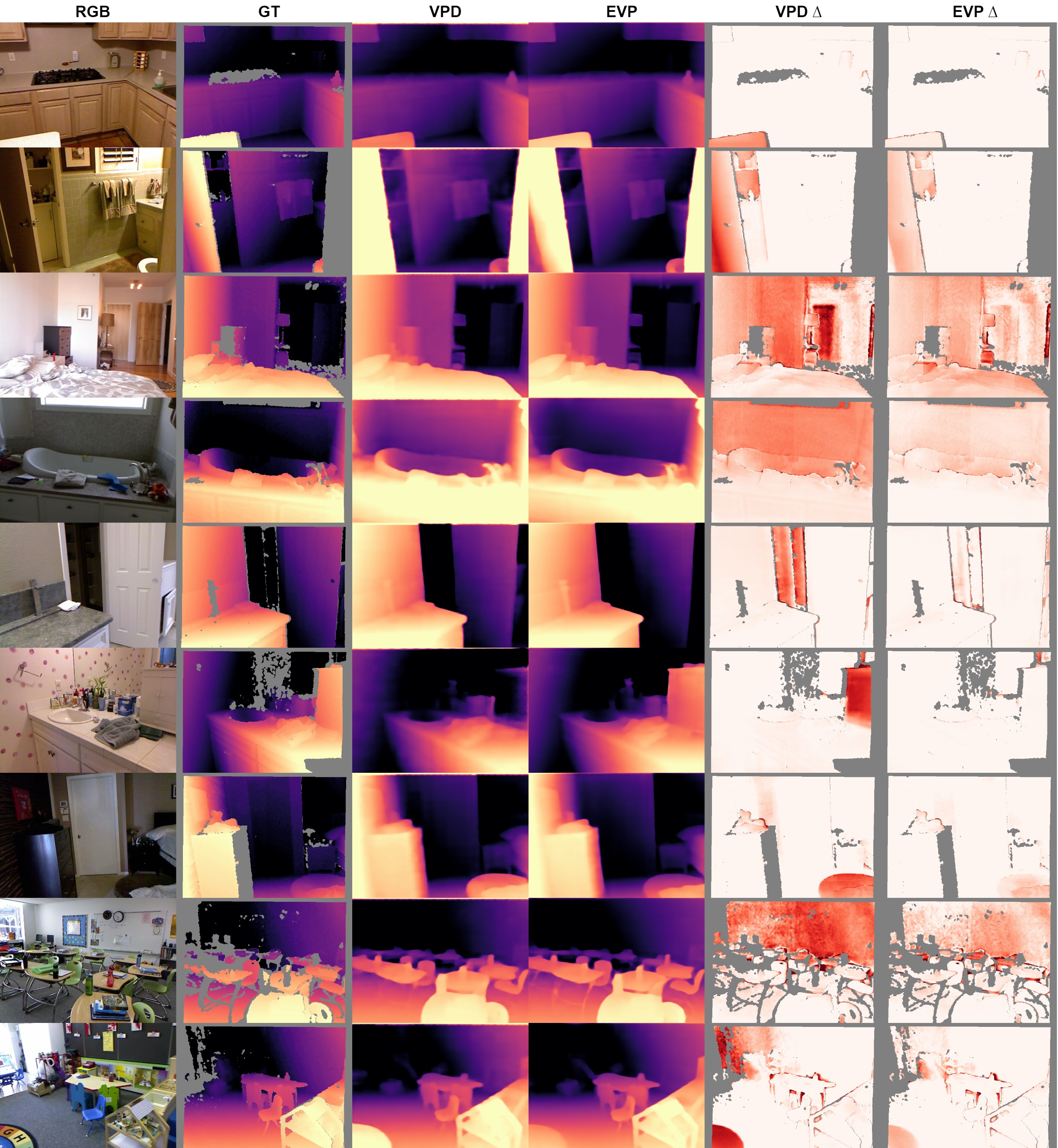}
   \hfill
   \caption{Visualization of EVP on images from the NYU Depth v2 dataset.}
   \label{fig:supsubfig1}
\end{figure*}

\begin{figure*}[t]
  \hspace{0.5cm}
   \includegraphics[width=1\linewidth]{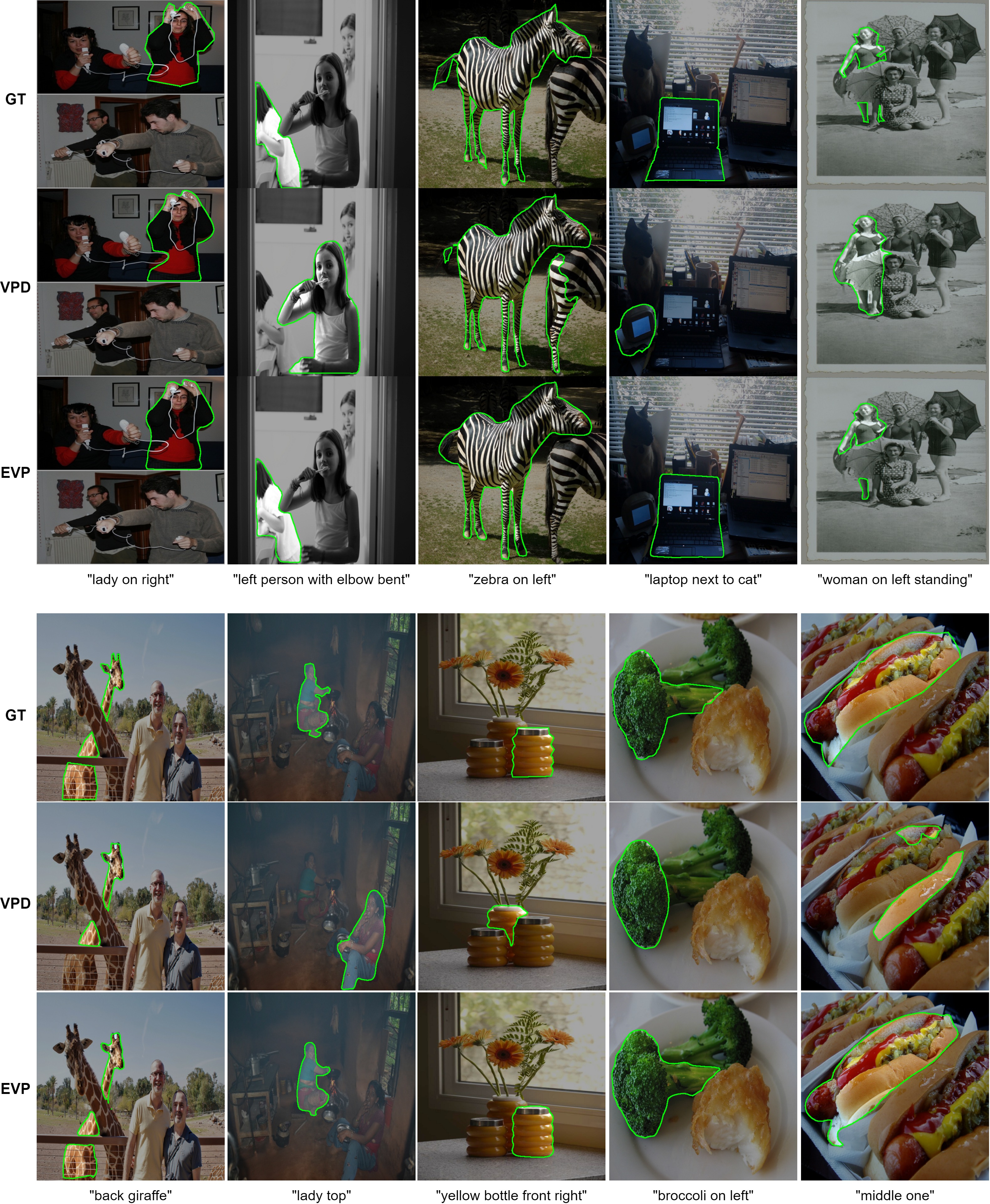}
   \hfill
   \caption{Visualization of EVP on images from the RefCOCO dataset.}
   \label{fig:supsubfig3}
\end{figure*}

\end{document}